\documentclass[twocolumn,a4paper]{article}
\usepackage[margin=3cm]{geometry}
\usepackage[utf8]{inputenc}
\usepackage{graphicx}
\usepackage{amsmath}
\usepackage{amssymb}
\usepackage{draftwatermark}
\usepackage[T1]{fontenc}
\usepackage{color}

\title{When Conventional machine learning meets neuromorphic engineering: Deep Temporal Networks (DTNets)  a machine learning frawmework allowing to operate on Events and Frames and implantable on Tensor Flow Like Hardware}
 \author{Marco Macanovic$^2$, Fabian Chersi$^2$, Felix Rutard$^2$, ,\\ , Sio-Hoi Ieng $^2$, Ryad Benosman$^{123}$\\ 
 ---------------------------------------------------------\\
$^1$ University of Pittsburgh/Medical Center, $^2$ Sorbonne Universitas,\\ ,\\ $^3$ Carnegie Mellon University,}

\begin{document}

\maketitle

\begin{abstract}

We introduce in this paper the principle of Deep Temporal Networks that allow to add time to convolutional networks by allowing deep integration principles not only using spatial information but also increasingly large temporal window. The concept can be used for conventional image inputs but also event based data. Although inspired by the architecture of brain that inegrates information over increasingly larger spatial but also temporal scales it can operate on conventional hardware using existing architectures.
We introduce preliminary results to show the efficiency of the method. More in-depth results and analysis will be reported soon! 

\end{abstract}

\section*{Introduction}

Temporal surfaces introduced in \cite{HOTS} allowed to learn temporal sequences of events on different temporal and spatial scales called HOTS (for Hierarchy of Temporal Surfaces). This learning method allows classification from the ouput of event based sensor, but can also be extended to any signal including conventional video sequences. Although promissing Neurmorphic event based engineering is still lacking a convincing computational paltform that allows to make full use of its potiential and mostly to take advantage of all the properties of event based cameras. HOTS is also very much linked to event based data and is not able to operate on frames and more important make use of existing GPU hardware.
Event-based  cameras are offering many advantages that frame-based cameras are not able to provide without an unreasonable increase in computational resources.  Low computation needs is especially achieved by a lower redundant data acquisition, while presenting lower latency than standard cameras, achieved via a  highly  precise  temporal  and  asynchronous  level  crossing  sampling.   
The main idea behind HOTS is to introduce a temporal context around incoming events to create a temporal context that we called a time surface. This concept can be extended to multiple time scale binding and integrating information on also larger sptial scales. 

The first layers of the architecture being fed with the output of the camera need to operate at time scales of a around hundred of microsecond to few millisecond that is incompatible with existing hardware.In fact altough processors can operate fast, random memory access as needed by sparse data output from event based camera is a major limitation specially when one want to operate fast at the native resolution of the sensor while being able to absorb event rates of the order of Giga Events per seconds.
Deeper scales however, are integrating information on larger times scales and do not need to operate fast. This is very similar to biological brains where low level areas need to process at the submillisecond the output of senses while higher areas and prefrontal cortex operate on slower time scales. We can then consider temporal integration steps above the order or 15-20 ms to be easily and efficiently performed by conventional architectures. Specially that when one integrates cues on larger time scales the amout of information tend to stabilize and decrease, the last layers are naturally compressing information.

AI has drastically improved during the last decade with the advent of more adapted hardware such as GPU with also the advent of  increasingly more optimized hardware. As we will show it is possible to take the best out of the conventional narrow AI and the neuromorphic world. The solution lays in rethinking temporal and spatial integration to be able to operate on GPU like hardware while maintaining its concept of temporal and spatial scale integration.  We coined this Methodology Deep Temporal Nets to be more in adequation with the current trend of deep AI technolgy to which it can add a better use of time. 
DTNets rely on a new hierachical architecture relying on the use of autoencoders to compress Time surfaces. We can then also add pooling possiblities to full comply with existing deep structures.

The DTNets machine learning architecture is the same line of thought as \textbf{deep neural network}. 
It is based on two main computational features that constitute the complete architecture: the \textit{autoencoders} for features extraction and the \textit{classifiers} to determine classes.
DTNets take as input any type of asynchronous signals based on existing techniques to sparsify signals such as sigma-delta, level crossing or relative change. But is can also use conventional video sequences not requiring to encode graylevel values in time by generating 'fake' events.
DTNets are hierarchical multilayer structures composed of two computational blocks. The first block extracts features using autoencoders. Autoencoders allow to extract features and abstract them into more and more complex representation. The second block is a classification layer that takes as input the output from the last layer and feeds it into a classifier. \\
A wide variety of classifiers can be used, from Multi Layer Perceptron currenlty in use, SVM, to more advanced techniques such as LSTM. Currently the retained classifier is MLP based and uses backpropagation to train. It important to emphasize that both layers rely on backpropagation for feature extraction and classification.

\section{Time Surfaces}
\begin{figure}[h!]
  \centering
  \includegraphics[width=\columnwidth]{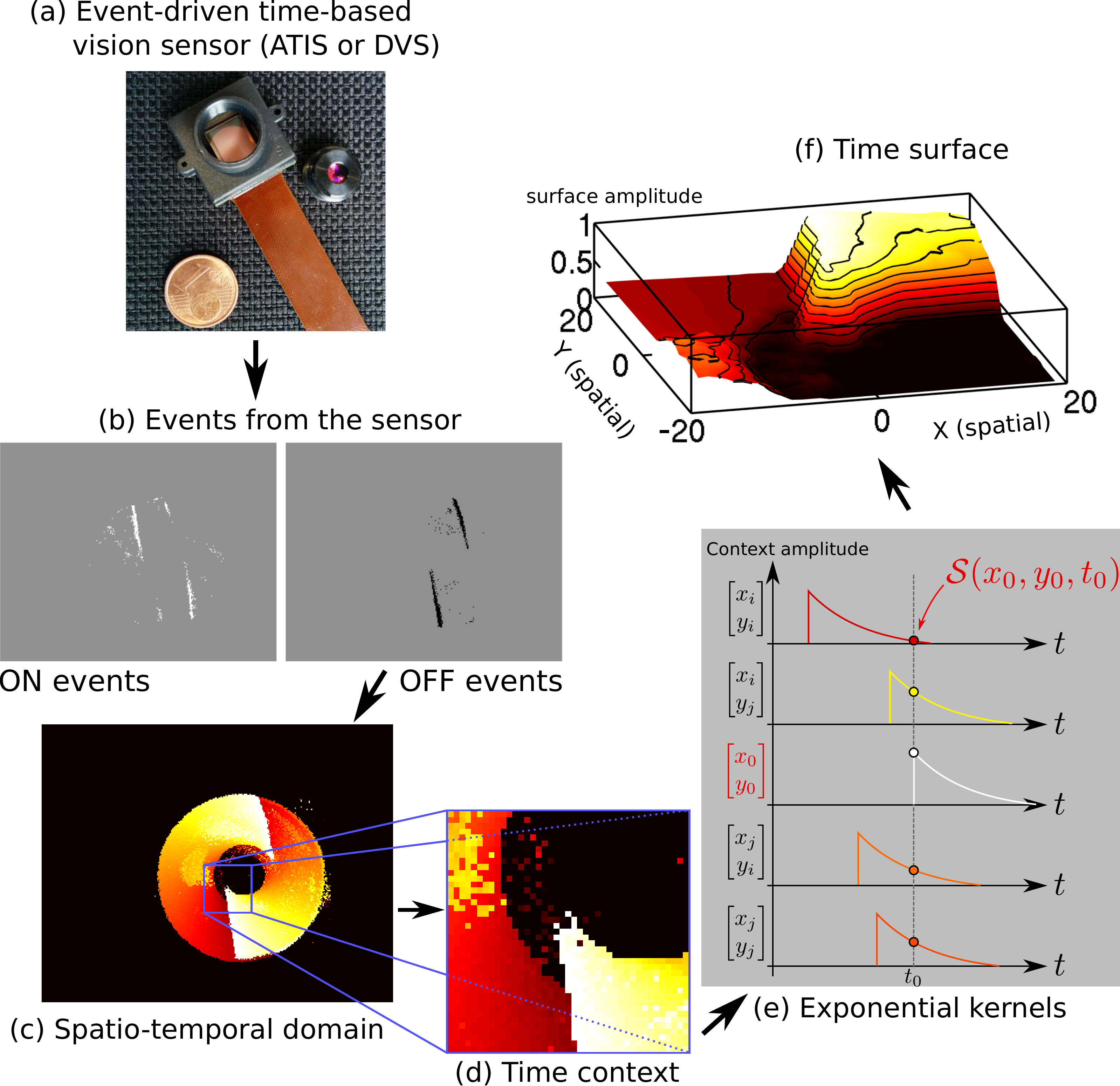}
  \caption{Definition of a time-surface from the spatio-temporal cloud of events from \cite{HOTS}.
           A time-surface describes the recent time history of events in the
           spatial neighborhood of an event. This figure shows how the time-surface
           for an event happening at pixel $\mathbf{x_0}=[x_0,y_0]^T$ at time $t_0$ is computed. The event-driven time-based vision sensor (a) is
           filming a scene and outputs events shown in (b) where ON events are represented on the left hand picture and
           OFF events on the right hand one. For clarity, we continue by only showing values associated to OFF events.
           When an OFF event $ev_i=[\mathbf{x_0},t_i,-1]$ arrives, we consider the times of most recent OFF events in the spatial neighborhood (c) where brighter
           pixels represent more recent events. Extracting a spatial receptive field 
           allows to build the event-context $\mathcal{T}_i(\mathbf{x},p)$ (d) associated with that event. Exponential decay kernels are then applyed to
           the obtained values (e) and their values at $t_i$ constitute
           the time-surface itself. (f) shows these values as a surface. This
           representation will be used in the following figures and the label of the axes will
           be removed for better clarity.
           }
  \label{fig:spike_context_def}
\end{figure}

\label{subsec:event_context}
The process of building a time-surface from the output of an event-driven time-based vision sensor as introduced in \cite{HOTS}
is illustrated in Fig.~\ref{fig:spike_context_def} and described hereafter.

Consider a stream of visual events (Fig.~\ref{fig:spike_context_def}(b)) which can be mathematically defined as
\begin{equation}
\begin{array}{l r}
ev_i = [\mathbf{x_i}, t_i, p_i]^T, & i \in \mathbb{N}\\
\end{array}
\end{equation}
where $ev_i$ is the $i^{th}$ event and consists of a location ($\mathbf{x_i}=[x_i, y_i]^T$), time ($t_i$) and polarity
($p_i$), with $p_i \in \{-1,1\}$, where $-1$ and $1$ represent OFF and
ON events respectively.
When an object (or the camera) moves, the pixels asynchronously
generate events which form a spatio-temporal point cloud representing the object's
spatial distribution and dynamical behavior. Fig.~\ref{fig:spike_context_def}(b) shows such
events generated by an object rotating in front of the sensor ((Fig.~\ref{fig:spike_context_def}(a)) where
ON and OFF events are represented respectively by white and black dots.

Because the structure of this point cloud contains information about the
object and its movement, we introduce the time-surface
$\mathcal{S}_i$ of the $i^{th}$ event $ev_i$ to keep track of the activity surrounding the spatial location $\mathbf{x_i}$ just before time $t_i$.
We can then define $\mathcal{T}_i(\mathbf{u},p)$ a time-context around an incoming event $ev_i$ as the array of most recent
events times at $t_i$ for the pixels in the $(2R+1)\times(2R+1)$ square neighborhood centered at $\mathbf{x_i}=[x_i,y_i]^T$ as:

\begin{equation}
\mathcal{T}_i(\mathbf{u},p) = \max_{j \leq i} \left\{t_j\,|\, \mathbf{x_j}=(\mathbf{x_i}+\mathbf{u}),\,p_j=p\right\},
\end{equation}
where $\mathbf{u}=[u_x,u_y]^T$ is such that $u_x \in \{-R,\dots,R\},$ $u_y \in \{-R,\dots,R\}$ and $p \in \{-1,1\}$
$\mathcal{T}_i(\mathbf{x},p)$ is shown in Fig.~\ref{fig:spike_context_def}(d) where intensity
is coding for time values: bright pixels show recent activity whereas dark pixels received events
further away in the past (only time values corresponding to OFF events are represented in the figure
for clarity).

Let $\mathcal{S}_i(\mathbf{u},p)$ be the time-surface around an event $ev_i$ (shown in Fig.~\ref{fig:spike_context_def}(e)), it is defined by applying an exponential decay kernel with time
constant $\tau$ on the values of $\mathcal{T}_i(\mathbf{u},p)$.
\begin{equation}\label{eq:context}
\mathcal{S}_i(\mathbf{u},p) =  e^{-(t_i-\mathcal{T}_i(\mathbf{u},p))/\tau}.
\end{equation}
$\mathcal{S}_i$ provides a dynamic spatiotemporal context around an event, the exponential decay expands the activity of passed events and provides information about the history of the activity in the neighborhood.
The resulting surface $\mathcal{S}_i(\mathbf{u},p)$ is shown in Fig.~\ref{fig:spike_context_def}(f)
for the OFF events represented all along Fig.~\ref{fig:spike_context_def}. In the following sections $\mathcal{S}_i(\mathbf{u},p)$ will be referred to directly as $\mathcal{S}_i$ to simplify notations. In the figures it will be represented as a surface showing the values of each of its element at their corresponding spatial positions.\\

\section*{Hierarchy of Temporal and Spatial Integration}
\subsection{Creating a Hierarchical Model}
\label{subsec:hierarchy_event_context}
\begin{figure*}
    \centering
    \includegraphics[width=\textwidth]{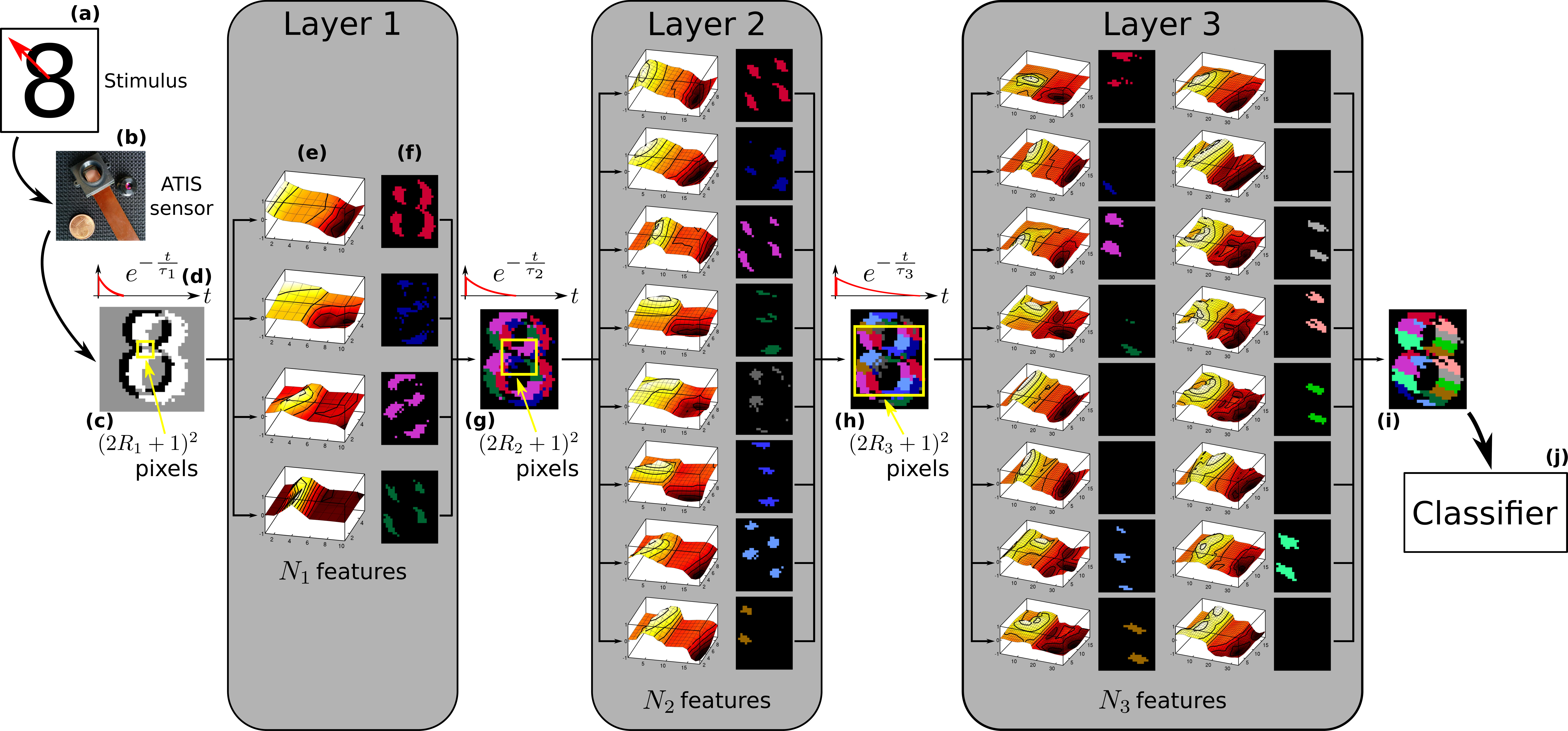}
    \caption{View of the proposed hierarchical model as introduced in \cite{HOTS}. From left to right, a moving digit (a) is presented to the event based camera (b) which produces ON and OFF events (c) which are fed into Layer $1$.
             The events are convolved with exponential kernels (d) to build event contexts from spatial receptive field
             of sidelength $(2R_1+1)$. These contexts are clustered into $N_1$ features (e). When a feature is matched, it
             produces an event (f). Events from the $N_1$ features are merged into the output of the layer (g).
             Each layer $k$ (gray boxes) takes input from its previous layer and feeds the next by reproducing
             steps (d)-(g). The output of Layer $k$ is presented between Layer $k$ and $k+1$ ((g),(h),(i)). To compute event contexts,
             each layer considers a receptive field of sidelength $(2R_k+1)$ around each pixel. The event
             contexts are compared to the different features (represented as surfaces in the gray boxes) and
             the closest one is matched. The images next to each features show the activation of their
             associated features in each layer. These activations are merged to obtain the output of the layer.
             The output (i) of the last layer is then fed to the classifier (j) which will recognize the object.}
    \label{fig:architecture}
\end{figure*}

Fig.~\ref{fig:architecture} illustrates the hierarchical model we introduce in this paper.
Steps (a) to (g) sum up the process described in the previous sections. As shown in Fig.~\ref{fig:architecture}, a moving digit (a) is presented
to the event based camera (b) which produces ON and OFF events (c). Time-surfaces are built by convolving
them with an exponential kernel of time constant $\tau_1$ (d) and considering spatial receptive fields of sidelength $(2R_1+1)$.
These time-surfaces are then clustered into $N_1$ prototypes represented as surfaces (e) in the Layer $1$ box.
When a cluster center is matched, an event is produced, resulting in the activations shown in (f). These events are
merged to form the output of Layer $1$ (g). One can see that each incoming event from the observed pattern is associated with the most representative prototype surface.\\
The nature of the output of Layer $1$ is exactly the same as its input: Layer $1$ outputs timed events. Once a prototype matches the temporal surface around the incoming event it immediately emits an event. Thus, the same steps used in Layer 1 (from (d) to (g)) can be applied in Layer $2$. However the emitted event is now representing the temporal activity of a prototype surface, it thus carries more meaning than the initial camera event. The prototype surfaces of Layer $2$ represent the temporal signature of the activity of complex features. Layer 2 uses different constants for space-time integration of features ($R_2$, $N_2$ and $\tau_2$). The goal is to introduce stability of the perceptual representation and sensitivity to the accumulation of sensory evidence over time. This integration over longer and longer time period will thus be able to accumulate evidence in favor of alternative propositions in a recognition process.
When alternatives with a barely discernible difference in their sensory inputs are presented over an extended period of time, longer time and spatial integration scales can accumulate the small differences over time until it becomes eventually possible to discriminate the alternatives through its ever growing output. This accumulation dynamics is at the heart of the HOTS model, the difference between time scales can be substantial and can start from 50ms for Layer $1$ to 250ms for Layer $2$ to finally reach 1.25 s for Layer $3$. \\
Layer $3$ receives input from Layer $2$, it is the last layer of the system and it provides the highest level information integration, as shown in Fig.~\ref{fig:architecture}(i) time-surface prototypes are also larger both spatially and temporally. The output of the temporal activity of Layer $3$ can finally be used for object recognition by being fed to a classifier (shown in Fig.~\ref{fig:architecture}(j)).

As stated above, each layer is then defined by only a few parameters (we add an index $l$ for the $l$th layer of the system):
\begin{itemize}
  \item $R_l,$ which defines the size of the time-surface neighborhood
  \item $\tau_l,$ the time constant of the exponential kernel applied to events
  \item $N_l,$ the number of cluster centers (prototypes) learnt by the clustering algorithm.
\end{itemize}

To increase the information extracted by each subsequent layer, we make these parameters
evolve between subsequent layer. For each layer, we define the parameters $K_R$, $K_\tau$, $K_N$
so that:
\begin{eqnarray}
  R_{l+1} &=& K_R\cdot R_l \\
  \tau_{l+1} &=& K_\tau\cdot\tau_l \\
  N_{l+1} &=& K_N \cdot N_l
\end{eqnarray}

The obtained architecture consists in a Hierarchy Of Time-Surfaces (HOTS) which is building and
extracting a set of features (the prototypes from the final layer) out of a stream of input events.
The time-surface prototypes will then be called time-surface features in the rest of the paper.

\subsection{Replacing Clustering by Auto-Encoders, compressive coding and more}

\begin{figure}[t!]
  \centering
  \includegraphics[height=.65\columnwidth]{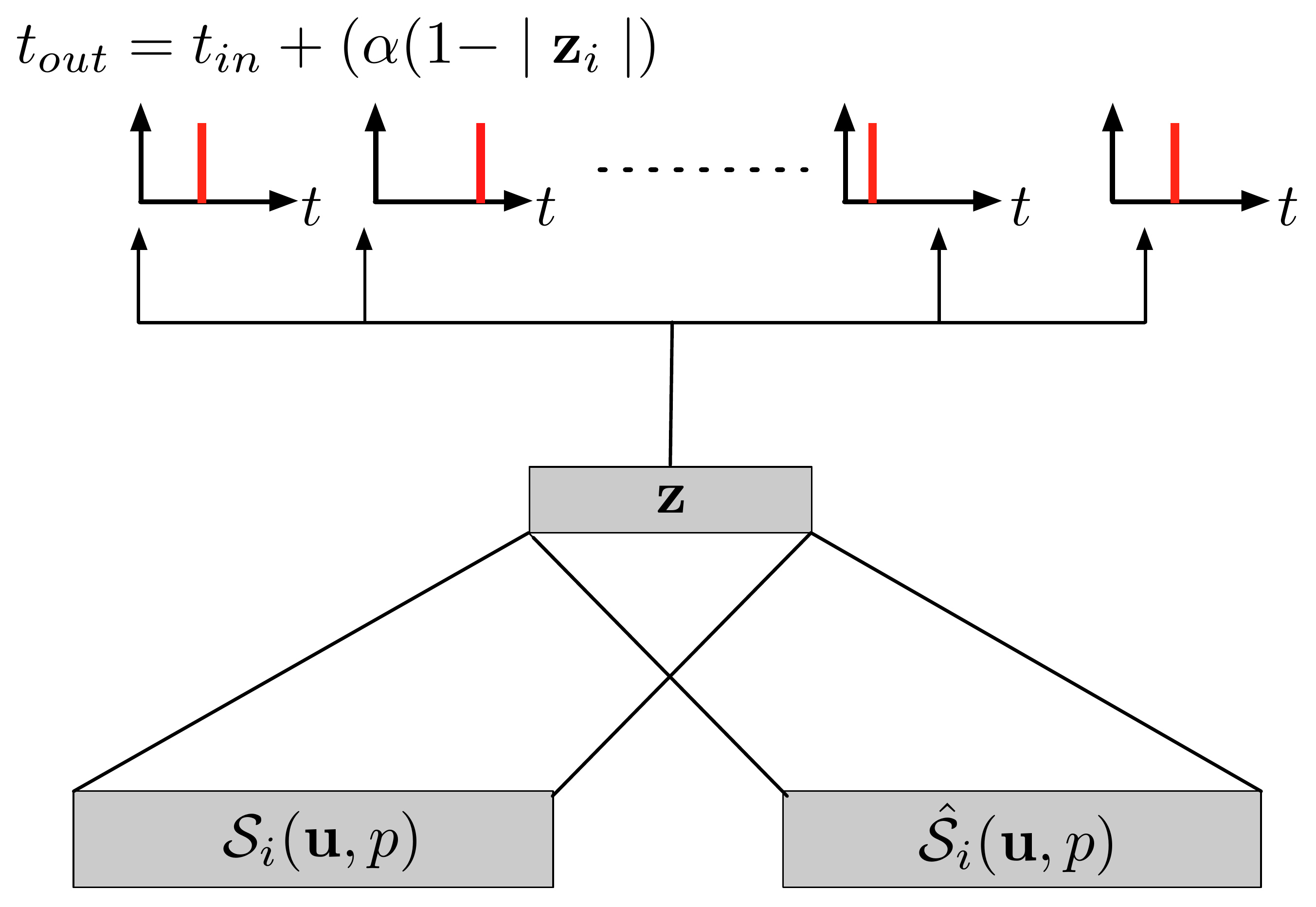}
  \caption{Schematics of the single layer autoencoder used the select features from incoming time surfaces.}
  \label{autoencoder}
\end{figure}

Clustering has been used to determine from a set of computed temporal contexts which ones are the most used, namely representing dominant information of scenes. Several other techniques could be used according to what type of information one would want to extract. In \cite{compressivehots} we used a sparse coding approch such as introduced in \cite{olshausen96}.
One could consider many other strategies to extract valuable time surfaces. Here we use autoencoders as compressive mean of selecting time surfaces.
This has the advantage of providing a compact representation of features and more important a higher generalization of features.
An autoencoder always consists of two parts, the encoder and the decoder (see Fig.\ref{autoencoder}), which can be defined as transitions $\phi$,$\psi$ such that:

$$\phi : \mathcal{X}\rightarrow\mathcal{F}$$
$$\psi :\mathcal {F}\rightarrow \mathcal {X}$$
$$ \phi ,\psi =\arg \min _{\phi ,\psi }\|\hat{\mathcal{S}}_i(\mathbf{u},p)-(\psi \circ \phi )\mathcal{S}_i(\mathbf{u},p)\|^{2}$$

We consider a single hidden layer, the encoder stage of the autoencoder takes the input volume patch $\mathcal{S}_i(\mathbf{u},p),$ and maps it to $\mathbf{z}\in\mathbb{R}^{p}=\mathcal{F}$ :

    $$\mathbf{z} =\sigma (\mathbf{W\mathcal{S}_i(\mathbf{u},p)} +\mathbf{b})$$

$\mathbf {z}$ is usually referred to as code, latent variables, or latent representation. Here, $\sigma$ is an element-wise activation function such as a sigmoid function or a rectified linear unit.

$\mathbf{W}$ is a weight matrix and $\mathbf {b}$ is a bias vector. After that, the decoder stage of the autoencoder maps $\mathbf {z}$ to the reconstruction $\hat{\mathcal{S}}_i(\mathbf{u},p)$ of the same shape as :

   $$ \hat{\mathcal{S}}_i(\mathbf{u},p)=\sigma '(\mathbf {W'z} +\mathbf {b'} )$$

where $\mathbf {\sigma'},\mathbf {W'} ,\text{ and } \mathbf {b'} $ for the decoder may differ in general from the corresponding $\mathbf {\sigma } ,\mathbf {W} ,{\text{ and }}\mathbf {b} $ for the encoder, depending on the design of the autoencoder. 

The hidden layer $\mathbf{z}$ encode a compressed version of time surfaces and can be used as an output feature to the next layer. Several possiblities can considered. The content can be sent as it is to the next layer. One can also consider as introduced in \cite{compressivehots} to encode values of $\mathbf{z}$ as  time delay, as shown in Figure.\ref{autoencoder} by considering the new time as the time of arrival of the time surface $t_{in}$ plus an additional delay with regards to the value of value of $\mathbf{z}_i$ weighted by a scalar $\alpha$.  One can also consider more crude version by simply thresholding the values of value of $\mathbf{z}_i$ to operate in the next layer on the meaninglfull components of $\mathbf{z}$. if raw values are used one could use a pooling layer similarly using a wide variety of pooling strategies similarly to what is being done in conventional deep convolution networks

The general architecture of the system is then as follows: as show in Figure.







\section{General Architecture}
\begin{figure*}[h!]
  \centering
  \includegraphics[width=\textwidth]{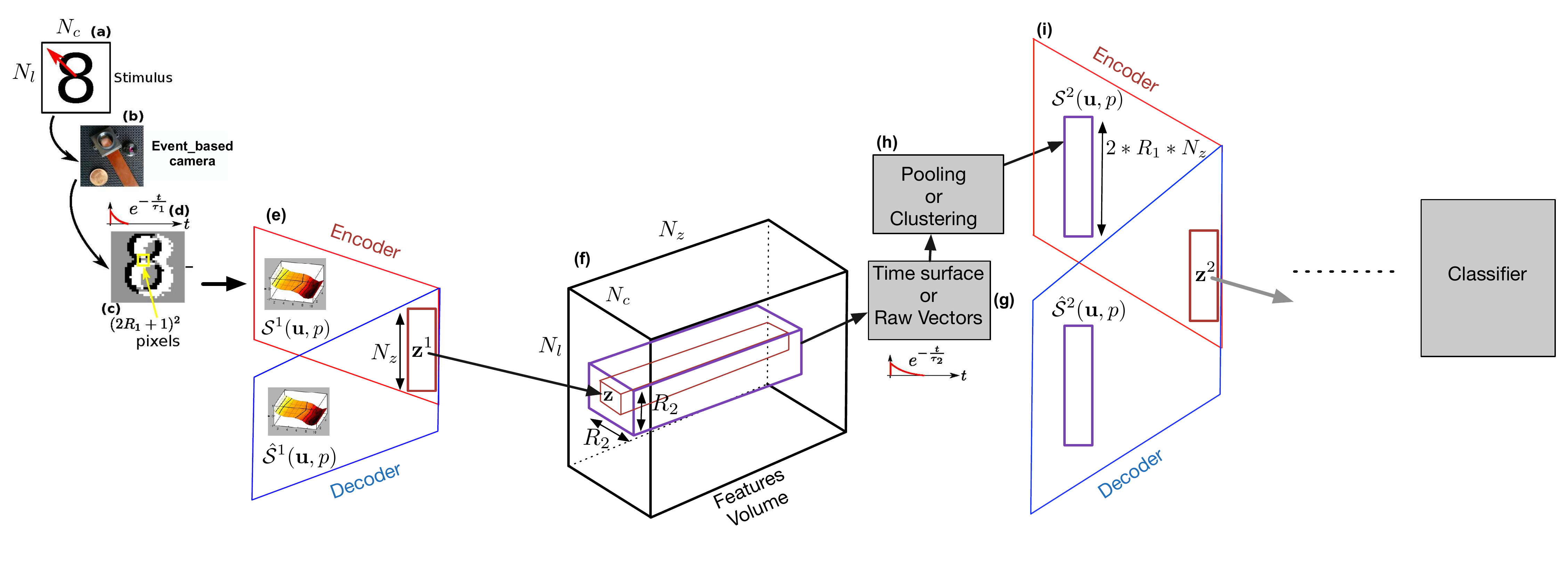}
  \caption{General architecture of DTNets.}
  \label{archidtnets}
\end{figure*}

A moving pattern observed by the event based camera (Figure.\ref{archidtnets}(a)). For each incoming event at a spatial location $\mathbf{p}$, we define a neighbourhood as shown in yellow in Figure.\ref{archidtnets}(c) for which we which we compute a time surface $\mathcal{S}^1(\mathbf{u},p)$ centered on the event (the subscript 1 stands for the layer 1). The time surface is fed into the autocoder that will compress the time surface and reduce dimensionality (\ref{archidtnets}(c)). Training is performed layer per layer, therefore at an initial stage, the autoencoder is trained for each incoming time surface, one the reconstruction erros is lower that a certain percentage, the training is considered complete, and the output of the encoder is sent to the rest of the processing chain. The autoencoder hidder layer $\mathbf{z}^1$ contains the compressed information and its size $N_z$ is smaller than the size of the input time surface $(2R_1+1)^2$. The output if the encoder is stored in a feature space (or features volume) at the location $\mathbf{p}$. The size of the volume is equal to the size of the focal plane $N_l*N_c$ (minus $2*R_1$ because of borders, but for clarity this precision is omitted in the figure), and depth $N_z$, the length of the hidden layer $z^1$.
We can then define a volume in the feature space of size $R_2*R_2*N_z$ as show in purple in Figure.\ref{archidtnets}(f). The content of the volume can processed in several ways. Many options are possible, the content of the volume can be kept as it is and pooled, or it can be encocded in the time domain on a larger time scale $\tau _2$, then clustered. All these strategies have both advantages and disadvantages that will be discussed further. The output of these two stage shown in Figure.\ref{archidtnets}(g-h)
is then sent to another time encoder that will compress the information. The content of the hidder layer $ \mathbf{z}^2$ can then be sent to another volume and so on.  Finally, the last layer sends its ouput to a classifier.

\section{Experiments}
we will first evluate a network's configuration that uses the direct ouput of the autoencoders wihtout adding any temporal concept. The network is simply encoding time surfaces at the inpout layer and then similarly to what conventional convolution networks operate. The idea of the network in to create feature exploiting the combinatory of the input without adding more temporal integration scales.The network is basically working of learning sequences without adding more temporal information into the network. 

The perfomances are evaluated on two learning tasks: 1) the recognition of asynchronous handwritten characters from the N-MNIST database, used by the neuromorphic community. 2) the recognition of the presence or absence of a car in an N-CARS database. Once the auto-encoders are running with the best reconstruction rate, we go out of the last layer to realize the final overall volume of the example and then vectorize it to finally send it to a classifier, here a Multi Layer Perceptron (MLP).

\begin{figure*}[h!]
  \centering
  \includegraphics[width=\textwidth]{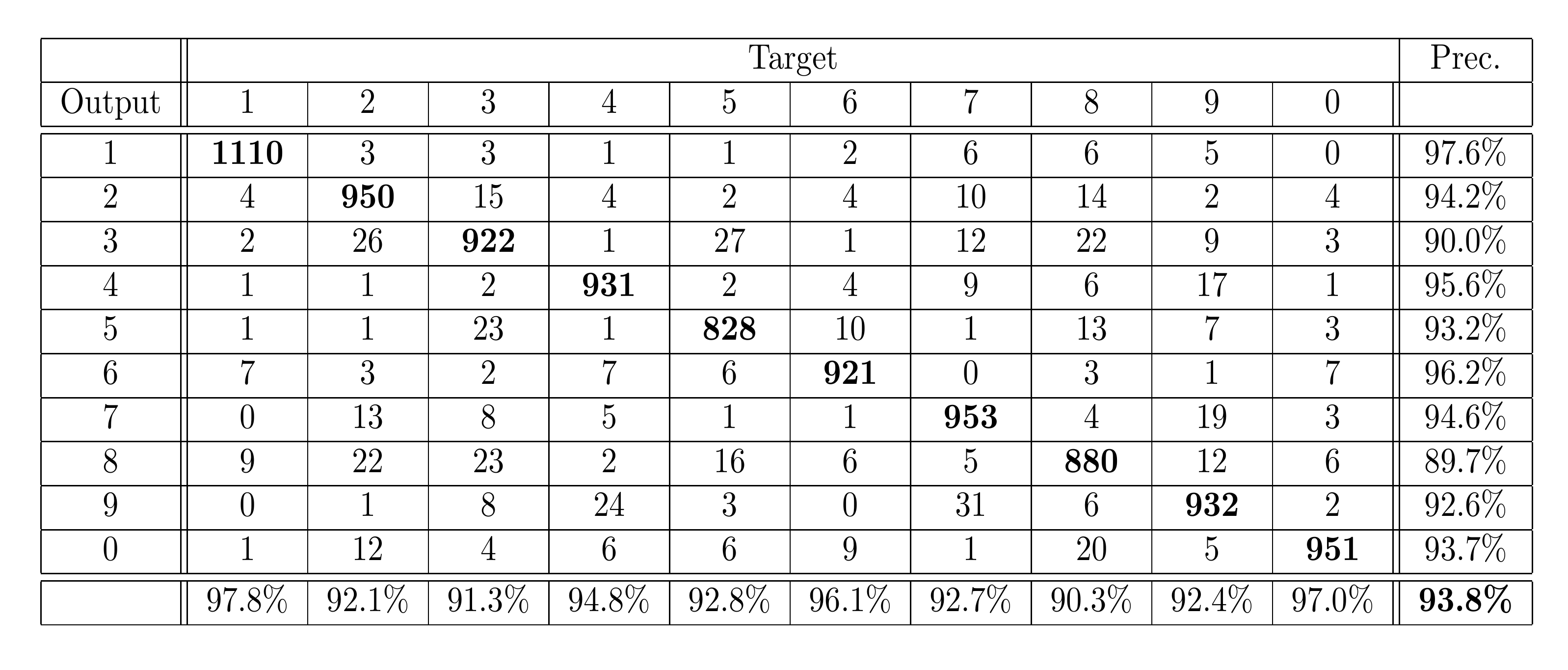}
  \caption{Confusion matrix on the NMNIST Database}
  \label{tab}
\end{figure*}

\begin{figure}[h!]
  \centering
  \includegraphics[width=.3\textwidth]{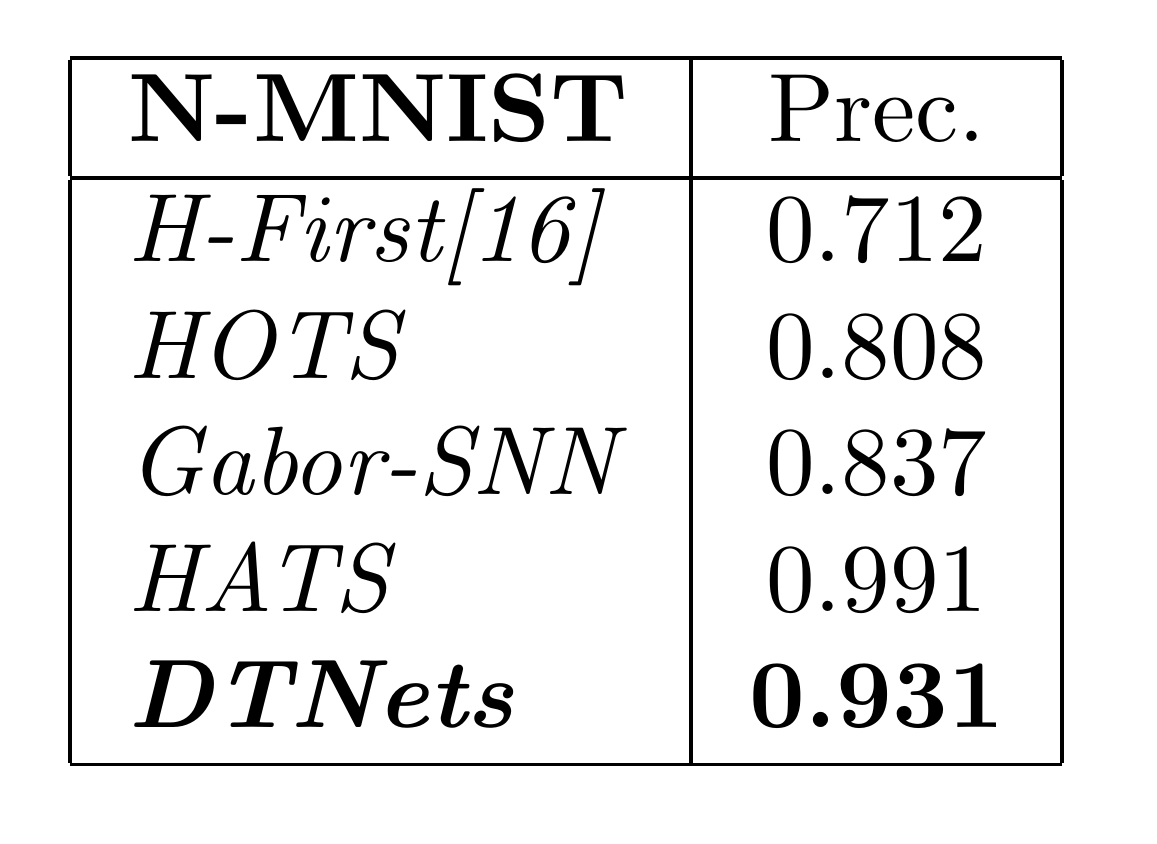}
  \caption{Recognition rates on N-MNIST}
  \label{tab2}
\end{figure}

We can see in Figure.\ref{tab} and Figure.\ref{tab2} that the overall recognition rate is 93.8\% based on N-MINIST data. The parameters used  are:
\begin{itemize}
\item Single layer of auto-encoder
\item A context radius $R_1$= 2
\item A time constant $\tau _2$ = 30ms
\item The number of neurons in the hidden layer of the auto-encoder N1 = 10 - The number of neurons in the hidden layer of the MLP NMLP = 200
\end{itemize}

\begin{figure}[h!]
  \centering
  \includegraphics[width=.3\textwidth]{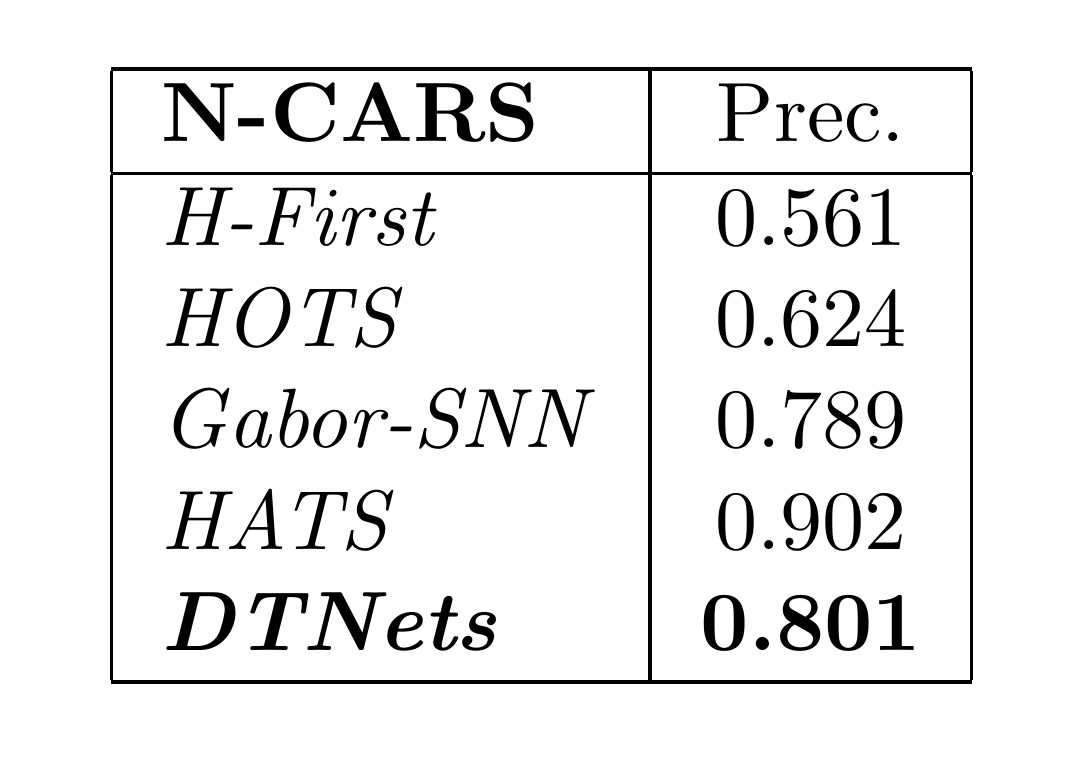}
  \caption{Recognition rates on N-CARS}
  \label{tab3}
\end{figure}

The N-CARS database is more complex to classify, we are introducing preliminary work, further tuning of the parameters will surely improve this rate (see Figure.\ref{tab2}). The parameters used are:
\begin{itemize}
\item Single layer of auto-encoder
\item A context radius $R_1$ = 3
\item A time constant $\tau _1$1 = 40ms
\item The number of neurons in the hidden layer of the auto-encoder N1 = 20
\item The number of neurons in the hidden layer of MLP NMLP = 200 The use of several layers greatly increases the dimensionality
\end{itemize}

\section{Conclusions}
These results although preliminary  shows that the algorithm performs well altough using only one layer and not integrating any additional temporal scales that is the essence of these temporal networks. We will report more in depth results and more advanced tuning strategies as soon as as our current non optimized implementation is improved as it currently tends to be slow to execute. 
Compared with other results of methods of classification of event based machine learning these Deep Temporal Networks provide promissing results and more imporant they can be easily implemented on GPU while maintaining an event based input but also any type of frame based input too. The method scales easily and can be added a variety of improvements to compress the data from layer to layer. We beleive that type type of structure not only integrates over space but can also integrzte easily over time while benefiting form the conventional machinel learning advances.

\section*{Aknowledgments}
We are gratefull to the fondation Voir et Entendre and also Streetlab to have funded this project. 
We also would like to thanks all member of the laboraotry not mentioned in this paper that also contributed to develop temporal networks.




\bibliographystyle{ieeetr}
\bibliography{DTNets}

\end{document}